\documentclass[11pt]{article}

\usepackage[final]{acl}

\usepackage{times}
\usepackage{latexsym}
\usepackage{orcidlink}

\usepackage[T1]{fontenc}
\usepackage[table]{xcolor}

\usepackage[utf8]{inputenc}

\usepackage{microtype}
\usepackage{subcaption} 

\usepackage{inconsolata}

\usepackage{graphicx}

\usepackage{listings}
\usepackage{xcolor}
\usepackage{float} 

\title{\textsc{VISHC at PsyDefDetect:}\\
Mitigating Data Scarcity in Psychological Defense Classification with Context-Aware Synthetic Augmentation}

\author{
 \textbf{Hoang-Thuy-Duong Vu\textsuperscript{1,2}\orcidlink{0009-0005-6775-6691}},
 \textbf{Quoc-Cuong Pham\textsuperscript{1,2}\orcidlink{0009-0005-6633-8201}},
 \textbf{Huy-Hieu Pham\textsuperscript{1,2,3}\orcidlink{0000-0003-4851-2518}\thanks{Corresponding author: \texttt{\href{mailto:hieu.ph@vinuni.edu.vn}{hieu.ph@vinuni.edu.vn}}}}
\\
\\
 \textsuperscript{1}College of Engineering and Computer Science, VinUniversity, Hanoi, Vietnam \\
 \textsuperscript{2}VinUni-Illinois Smart Health Center, VinUniversity, Hanoi, Vietnam \\
 \textsuperscript{3}Center for Innovations in Health Sciences, VinUniversity, Hanoi, Vietnam
\\
 \texttt{\{26duong.vht, 24cuong.pq, hieu.ph\}@vinuni.edu.vn}
}

\begin{document}
\maketitle

\begin{abstract}
Psychological defense mechanisms (PDMs) are unconscious cognitive processes that modulate how individuals perceive and respond to emotional distress. 
Automatically classifying PDMs from text is clinically valuable but severely hindered by data scarcity and class imbalance, challenges which generative augmentation alone cannot resolve without psychological grounding. 
In this work, we address these challenges in the PsyDefDetect shared task (BioNLP@ACL 2026) by proposing a context-aware synthetic augmentation framework combined with a hybrid classification model. 
%
%
Our hybrid model integrates contextual language representations with basic clinical features, along with 150 annotated defense items.
Experiments demonstrate that definition quality in prompting directly governs generation fidelity and downstream performance. 
Our method surpasses \textsc{DMRS Co-Pilot}, reaching an accuracy of 58.26\% (+40.25\%) and a macro-F1 of 24.62\% (+15.99\%), thereby establishing a strong baseline for psychologically grounded defense mechanism classification in low-resource settings.
%
Source code is available at: \href{https://github.com/htdgv/CASA-PDC}{https://github.com/htdgv/CASA-PDC}.

\end{abstract}

\section{Introduction}

Psychological Defense Mechanisms (PDMs) present a unique challenge for Natural Language Processing (NLP) field, particularly, they are unconscious, context dependent processes that appear through subtle cues such as narrative inconsistency, shifts in emotional framing, and distorted attribution, rather than clear lexical markers \cite{vaillant1994ego,cramer1987development,bond83}.
This implicit nature creates semantic ambiguity in which identical surface text may reflect distinct defensive processes, depending on underlying intent and psychological context, leading standard token or sentence level models to conflate adaptive coping with maladaptive defenses.

Data scarcity and class imbalance further compound these difficulties. 
Synthetic augmentation via Large Language Models (LLMs) offers a natural remedy, yet without psychologically grounded constraints, generative models produce fluent but theoretically invalid text, creating hallucinating defenses that introduce noise and 
erode model reliability \cite{Ji_2023,na-etal-2025-survey,anabytavor2019datadeeplearningrescue,kumar2021dataaugmentationusingpretrained}. 
A key point is that the PsyDefDetect shared task \cite{na-etal-2026-psydefdetect} on the \textsc{PsyDefConv} dataset \cite{na-etal-2026-psydefconv}, based on the \textsc{ESConv} dataset \cite{liu-etal-2021-towards}, introduces two auxiliary labels, \textit{No Defense} (Level 0) and \textit{Need More Information} (Level 8), that carry no corresponding clinical defense items \cite{diGiuseppePerry2021}.
These labels violate standard multi-class assumptions and produce skew distributions, making defense-item-based feature extraction underspecified, demanding a principled reformulation of the task.


    
    

We address these challenges in the PsyDefDetect shared-task through context-aware synthetic augmentation
paired with a Hybrid Feature Fusion architecture. Specifically, our contributions are:

\begin{itemize}
    \item \textbf{Psychologically grounded augmentation.} Stressor-anchored, theory-driven prompts with class-specific definitions from the Defense Mechanisms Rating Scales (DMRS) for synthetic augmentation to ensure generating high-fidelity examples, demonstrating that definition quality in prompting governs downstream performance.

    \item \textbf{Clinical feature engineering.} Structured features 
    from all 150 defense items, along with basic clinical features, are fused with contextual language representations, bridging clinical theory and neural classification.


    \item \textbf{Strong low-resource baseline.} Using Llama-3-8B-Instruct as data generator, our system improves accuracy (18.01\% to 58.26\%) and macro-F1 (8.63\% to 24.62\%) on the \textsc{PsyDefConv} blind-test set, establishing a competitive foundation for PDM classification.
\end{itemize}

\begin{figure}[t]
  \includegraphics[width=\columnwidth]{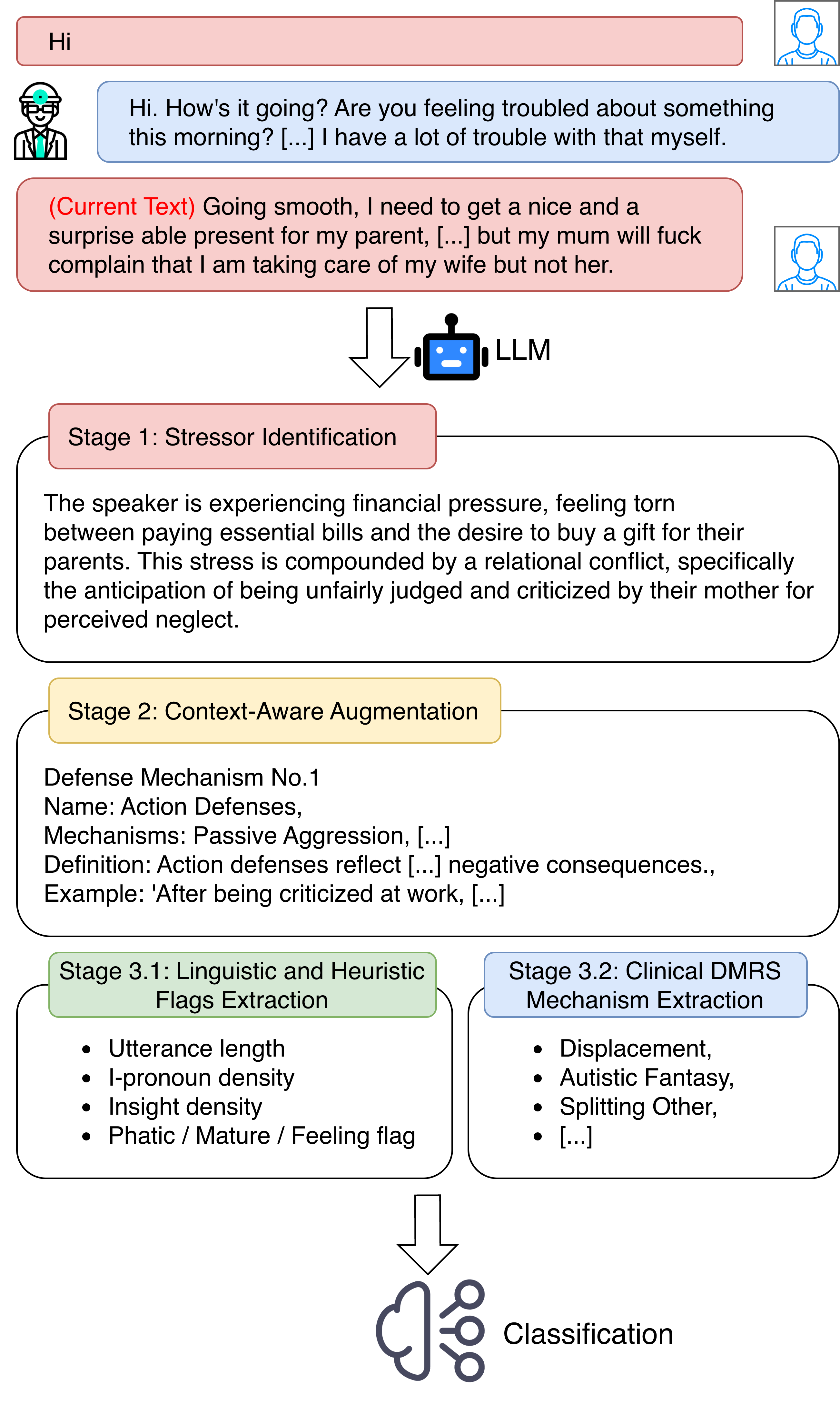}
  \caption{Overview of the multi-stage research pipeline. The process begins with (1) LLM-based stressor identification to establish contextual grounding; (2) context-aware synthetic data augmentation to address class imbalance; and (3) a dual-domain feature extraction stage targeting linguistic heuristics and clinical DMRS mechanism indicators; followed by the final classification.}
  \label{fig:architecture}
\end{figure}

\section{Methodology}
\label{sec:methods}

We address PDM classification under data scarcity through three stages:
(1) a context aware synthetic data augmentation pipeline grounded in clinical theory;
(2) a dual-domain feature extraction stage, combining linguistic heuristics with DMRS derived defense profiles; and
(3) a hybrid fusion architecture that integrates contextual language representations with structured clinical features (Figure~\ref{fig:architecture}).

\subsection{Context-Aware Data Augmentation}

Standard augmentation methods such as paraphrasing and back translation \cite{wei-zou-2019-eda} introduce diversity in phrasing but do not retain the functional role of defense mechanisms. We present a Synthetic Data Augmentation (SDA) pipeline based on Llama-3-8B-Instruct that captures the psychological conditions that give rise to defenses, with emphasis on function instead of form.

\paragraph{Stressor-Anchored Generation.}
Defense mechanisms emerge in response to perceived stressors rather than in isolation. Each prompt is anchored in a key stressor identified from the dialogue, such as interpersonal conflict, job loss, or social rejection. This approach promotes responses that reflect realistic defensive behavior instead of generic expressions of emotion.

\paragraph{Theory-Driven Prompting.}
To control semantic drift and reduce label inconsistency, each prompt defines the target defense level using structured clinical details from the DMRS framework \cite{diGiuseppePerry2021}. These details include the defense name, its formal definition, and common linguistic and behavioral patterns. Paired with few-shot examples (Appendix~\ref{appendix:prompt}), this design steers generation toward samples that express the theoretical role of each defense rather than relying on surface plausibility \cite{brown2020languagemodelsfewshotlearners}.

\subsection{Data Quality Control}

\begin{table}[t]
\centering
\small
\setlength{\tabcolsep}{6pt}
\begin{tabular}{llccc}
\hline
\textbf{Class} & \textbf{Label} & \textbf{N} &
\textbf{SB}$\downarrow$ & \textbf{SA}$\uparrow$ \\
\hline
No Defense     & 0 & 500 & 0.399 & 0.481 \\
Action         & 1 & 500 & 0.496 & 0.585 \\
Major Image    & 2 & 488 & 0.440 & 0.559 \\
Disavowal      & 3 & 500 & 0.453 & 0.597 \\
Minor Image    & 4 & 500 & 0.416 & 0.592 \\
Neurotic       & 5 & 384 & 0.392 & 0.525 \\
Obsessional    & 6 & 500 & 0.429 & \textbf{0.619} \\
Needs Info     & 8 & 224 & \textbf{0.413} & 0.601 \\
\hline
\textit{Avg.}  &   &     & \textbf{0.430} & \textbf{0.570} \\
\hline
\end{tabular}
\caption{Synthetic data quality per class. \textbf{SB}: Self-BLEU \cite{zhu2018texygenbenchmarkingplatformtext} (lower =
greater lexical diversity); \textbf{SA}: Semantic Adherence via Natural Language Inference (NLI) entailment
(higher = stronger label alignment). Class~7 (High-Adaptive) is excluded from
augmentation due to sufficient original samples; all other classes are capped
at $N{=}500$ synthetic instances.}
\label{tab:synthetic_quality}
\end{table}

Uncontrolled generation may introduce label noise and artifacts. To mitigate this, we use two quality control steps. First, a soft balancing scheme restricts each minority class to 500 total samples (real plus synthetic), which helps reduce overfitting to generation specific patterns. We examine five augmentation settings: $\times 1$, $\times 2$, $\times 5$, $\times 8$, and $\times 10$, together with the 500 cap variant. Second, a machine as annotator filter applies a secondary classifier to assign labels to generated batches; Only batches achieving a Cohen's Kappa of $\kappa \geq 0.60$ \cite{Cohen1960ACO}, reflecting substantial agreement, are retained (Table~\ref{tab:synthetic_quality}). 

\subsection{Feature Extraction}

Each seeker utterance is represented by two complementary feature sets: 
(i) lightweight linguistic heuristics capturing surface-level cues, and (ii) 
clinically grounded DMRS-derived features encoding latent defensive functioning.

\paragraph{Linguistic and Heuristic Features.}
We define six lightweight features to distinguish non-defensive (Label 0)
from defensive responses, which are often conflated: \textit{Utterance Length}
(narrative elaboration proxy), \textit{I-Pronoun Density} (self-focus),
\textit{Insight Density} (reflective reasoning), \textit{Phatic Flag} (short
filler utterances), \textit{Mature Coping Flag} (triggered by length $>12$,
high insight, and elevated I-pronouns), and \textit{Emotion Intensity} (model
confidence in non-neutral predictions).

\paragraph{DMRS Defense Profile.}
We approximate latent defensive functioning using a four-step indicator
inference procedure:
\begin{enumerate}
    \item \textit{Indicator Scoring:} An NLI model estimates entailment
    probability $P(T \Rightarrow I_j)$ for each of 150 DMRS indicators
    given utterance $T$.
    \item \textit{Mechanism Aggregation:} Indicator scores are grouped
    into 30 defense mechanisms and normalized to form mechanism scores $S(M_k)$.
    \item \textit{Profile Construction:} The resulting 30-dimensional
    vector defines the Defense Profile of the utterance.
    \item \textit{Level Mapping:} Mechanism scores are aggregated by DMRS
    level to obtain the predicted defense level:
    $\hat{y} = \arg\max_{\ell} \sum_{M_k \in \ell} S(M_k)$.
\end{enumerate}


\subsection{Hybrid Feature Fusion Architecture}

Our system integrates contextual language representations with structured clinical features using a late fusion approach, as described below:

\begin{enumerate}
    \item \textbf{Textual Encoder:} MentalRoBERTa \cite{ji2022mentalbert}
    encodes each instance formatted as \texttt{[Stressor:S|Turn:T]},
    conditioning the representation on both the triggering context and the
    response, yielding a 768-dim embedding.


    \item \textbf{Feature Encoders:} The heuristic (7-dim) and DMRS-derived (30-dim) features are each passed through a dedicated Multilayer Perceptron (MLP) with the following structure: 64 → Batch Normalization → ReLU → Dropout (p = 0.3) → 32, producing two 32-dim vectors.

    \item \textbf{Fusion and Classification:} The three representations
    are concatenated into an 832-dim vector
    (768 + 32 + 32) and passed through two fully connected layers
    (256 $\to$ 128, ReLU + Dropout ($p$=0.4)) and a final linear layer
    producing a probability distribution over 9 labels \cite{kiela2020supervisedmultimodalbitransformersclassifying}.
\end{enumerate}

\section{Experiment \& Results}


\label{sec:experiment}




\subsection{Experimental Setup}
\label{sec:setup}

\paragraph{Data Pre-processing.}
The training corpus combines human-annotated dialogues with synthetic samples
generated by Llama-3-8B-Instruct under theory-driven prompting. We evaluate
six augmentation scales, $\times$1, $\times$2, $\times$5, $\times$8,
$\times$10, and a hard cap of $N{=}500$ per class, yielding corpora ranging
from approximately 1,800 to 5,100 instances. Class 7 is excluded from
augmentation given its already substantial representation. The baseline
corresponds to the organizer-provided code rerun with Llama-3-8B-Instruct
and no augmentation. \footnote{Per-class metrics and DMRS activation patterns are detailed in Appendix~\ref{sec:appendix}.}

\paragraph{Implementation Details.}
The model is implemented in PyTorch using the Hugging Face Transformers
library. MentalRoBERTa (\texttt{mental-roberta-base}) \cite{ji2022mentalbert} 
serves as the textual encoder; its parameters are fine-tuned end-to-end with a 
learning rate of $1{\times}10^{-6}$. Task-specific layers (MLPs, fusion head, 
and classifier) use a higher learning rate of $1{\times}10^{-4}$, optimized 
with AdamW \cite{loshchilov2019decoupledweightdecayregularization}. Training runs for up to 20 epochs 
with batch size 16, early stopping on validation macro-F1, weight decay of 
$1{\times}10^{-2}$, and label smoothing ($\epsilon = 0.1$) to mitigate noise 
introduced by synthetic samples.

\paragraph{Evaluation Metrics.}
We follow the same protocol employed in \citet{na-etal-2026-psydefdetect} and report all results on both the \textsc{PsyDefConv} development and blind test sets using macro-averaged Precision, Recall, and F1, as well as overall Accuracy, to account for class imbalance.

\subsection{Results Analysis}
\label{sec:results}

\begin{table}[t]
\centering
\small
\setlength{\tabcolsep}{5pt}
\begin{tabular}{lcccc}
\hline
\textbf{Setting} & \textbf{Acc}$\uparrow$ & \textbf{P}$\uparrow$ & \textbf{R}$\uparrow$ & \textbf{F1}$\uparrow$ \\
\hline
\multicolumn{5}{l}{\textit{Baseline}} \\
\textsc{DMRS Co-Pilot}$^*$ & 0.1801 & 0.1904 & 0.1715 & 0.0863 \\
\hline
\multicolumn{5}{l}{\textit{Our System}} \\
$\times1$  & 0.5508 & 0.2555 & 0.2601 & 0.2543 \\
$\times2$  & 0.5508 & 0.2789 & 0.2882 & \textbf{0.2799} \\
$\times5$  & 0.5487 & 0.2764 & 0.2821 & 0.2783 \\
\rowcolor{green!20}
$\times8$$^\dagger$ & \textbf{0.5826} & \textbf{0.2588} & \textbf{0.2503} & \textbf{0.2462} \\
$\times10$ & 0.5254 & 0.2237 & 0.2289 & 0.2238 \\
$N{=}500$  & 0.5275 & 0.2659 & 0.2654 & 0.2628 \\
\hline
\end{tabular}
\caption{Classification performance on the \textsc{PsyDefConv} official 
test set. $^*$Baseline rerun with Llama-3-8B-Instruct on the released 
test set; original results in \citealt{na-etal-2026-psydefconv} use a 
different backbone. $\times k$: each minority class expanded to $k$ times 
its original size. $N{=}500$: hard cap of 500 instances per class. 
Metrics are macro-averaged. $^\dagger$Official leaderboard submission; 
all other rows are post-hoc evaluations on the released test set. 
Best results per column in \textbf{bold}.}
\label{tab:results}
\end{table}

\paragraph{Classification performance across settings.}

Our submission ranked 13 out of 21 registered teams in the official evaluation. Table~\ref{tab:results} reports results across all six augmentation configurations on the official blind-test set. Every augmented variant substantially outperforms \textsc{DMRS Co-Pilot} in accuracy (+40.25 pp) and macro-F1 (+15.99 pp), confirming that theory-driven augmentation delivers robust gains over a prompt-only LLM baseline in this low-resource setting.
Performance improves at lower augmentation scales but deteriorates as augmentation becomes more aggressive. The $\times 2$ configuration yields the highest macro-F1 (27.99\%), indicating an effective balance between expanded class coverage and synthetic generation noise. Further scaling leads to a steady decline in macro-F1, which falls to 22.38\% at $\times 10$, consistent with noise accumulation in heavily augmented corpora \cite{kumar2021dataaugmentationusingpretrained}. While $\times 8$ records the highest accuracy (58.26\%), its macro-F1 remains 3.37 pp below $\times 2$, revealing that overall accuracy is disproportionately influenced by dominant Label~7 predictions at the expense of minority-class recall.

\begin{figure}[t]
  \centering
  \begin{subfigure}{\columnwidth}
    \centering
    \includegraphics[width=\linewidth]{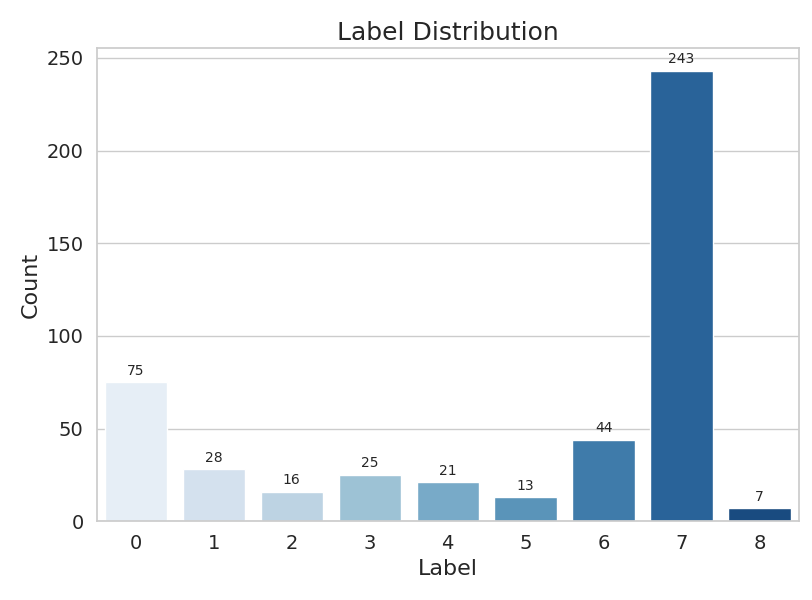}
    \caption{Label distribution of the \textsc{PsyDefConv} official test set.}
  \end{subfigure}
  \vspace{0.5em}
  \begin{subfigure}{\columnwidth}
    \centering
    \includegraphics[width=\linewidth]{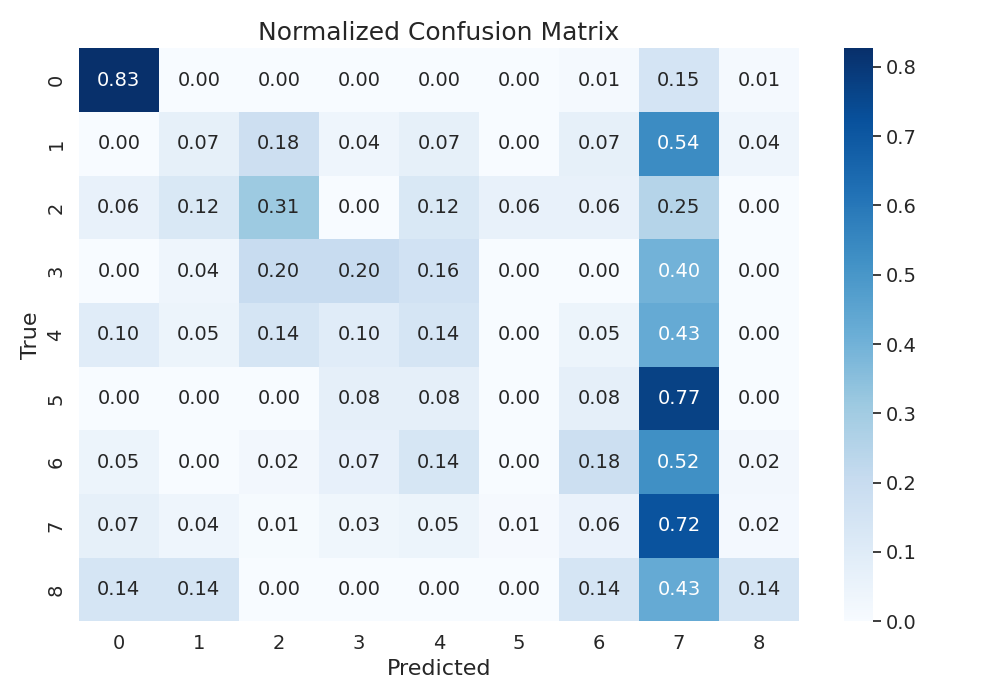}
    \caption{Row-normalized confusion matrix of our official leaderboard submission (\textsc{PsyDefConv} test set, $\times$8).}
  \end{subfigure}
  \caption{(a) The \textsc{PsyDefConv} official test set label distribution and (b) row-normalized confusion matrix of our official leaderboard submission ($\times$8). Label~7 dominates both the distribution (243/472 instances) and predictions, absorbing errors from all other classes.}
  \label{fig:cm_normalized}
\end{figure}

\paragraph{The Label~7 sink effect and class imbalance.}
The confusion matrix (Figure~\ref{fig:cm_normalized}b) confirms Label~7 as a 
universal prediction sink. The most severe case is Label~5 (Neurotic), where 
77\% of instances are misclassified as Label~7, rendering this class 
unlearnable. Per-class F1 scores shows that Labels~0 and~7 exceed 
F1~$>$~0.70, while all remaining classes fall below 0.30, with four classes 
below 0.15. This implies that the accuracy (0.55-0.58) substantially overstates 
practical utility. 
The Label~5/7 confusion is semantically meaningful: both involve reflective discourse, but differ in whether anxiety is intellectualized or channeled.

\paragraph{The primacy of definition quality.}
A key finding is the sensitivity of model performance to prompt design. Compared to the baseline of \citet{na-etal-2026-psydefconv}, which uses shallow class descriptions and achieves a Macro-F1 of 8.63\%, our best setting ($\times2$) improves by 19.36 pp. We attribute this to the richer definitional context provided by our DMRS-based definitions derived from \citet{diGiuseppePerry2021}, which better disambiguate overlapping classes and reduce label ambiguity during generation.
\section{Conclusion}

Clinical specificity of class definitions is the primary driver of synthetic augmentation effectiveness in PDM classification. Our hybrid system, combining MentalRoBERTa with DMRS-derived features and stressor-anchored generation, achieves substantial gains over \textsc{DMRS Co-Pilot} (Accuracy: 18.01\%$\to$58.26\%; macro-F1: 8.63\%$\to$24.62\%). However, the Label~7 sink effect and resulting bimodality indicate that augmentation alone cannot overcome majority-class bias and clinical proximity. 
Future works should consider including constraint-based decoding, human-in-the-loop validation, and dialogue-level modeling to address temporal volatility identified in our analysis.

\section*{Limitations}

\paragraph{Majority-class dominance and augmentation ceiling.}
The most critical limitation is the Label~7 sink effect identified in Section~\ref{sec:results}. Despite augmenting minority classes to $N = 500$, the model's decision boundary remains heavily biased toward Label~7 (243/472 development instances), and macro-F1 performance on six of eight classes remains below 0.30. This suggests that naive count-balancing is insufficient when the majority class also exhibits high linguistic surface overlap with adjacent classes. Addressing this will require loss re-weighting strategies (e.g., focal loss \cite{lin2018focallossdenseobject}), hard-negative mining during augmentation, or explicit contrastive learning objectives that sharpen inter-class boundaries rather than simply expanding minority class size.

\paragraph{Turn-level modeling and temporal blindness.}
Our proposed architecture operates on isolated seeker utterances, each formatted with only the preceding stressor context. However, our dataset analysis (Appendix~\ref{sec:appendix_dataset}) demonstrates that defense levels are unstable, frequently undergoing abrupt transitions across turns, and that larger defensive shifts tend to occur rapidly (Figure~\ref{fig:correlation_magnitude_speed}). A turn-level classifier observes only the outcome of a defensive transition, not the trajectory that produced it. This architectural limitation is especially problematic for clinically adjacent classes (e.g., Labels~6 and~7), whose distinction may reside in discourse-level patterns spanning multiple turns rather than in any single utterance.

\paragraph{Synthetic data validity and clinical reliability.}
Our quality control pipeline filters for inter-annotator agreement ($\kappa \geq 0.60$) using a secondary classifier, providing a practical proxy for label consistency. However, this process does not guarantee clinical validity. A generated utterance may receive consistent classification by both the primary LLM and the secondary classifier while still failing to instantiate the functional psychological role of the target defense. Without human expert validation of a representative sample of synthetic instances, we cannot bound the rate of theoretically invalid but classifier-plausible samples in our training data. This is a fundamental limitation of machine-as-annotator pipelines in psychologically grounded domains, and future work should incorporate systematic clinician review.


\paragraph{Label underspecification and task formulation ambiguity.}
Labels~0 (No Defense) and~8 (Needs More Information) do not correspond to clinical defense mechanisms and thus lack the DMRS indicator structure used to construct our Defense Profile features. For Label~8, the classification signal must be derived from the textual encoder and heuristic features, with the DMRS branch contributing noise rather than discriminative signal. Meanwhile, Label~0 is defined by the absence of a positive mechanism, making it harder to synthesize and harder for the NLI-based indicator scoring to characterize. A principled resolution, treating Labels~0 and~8 as a prior detection stage (defensive and non-defensive and ambiguous) before running the eight-class classifier, is deferred to future work.


\section*{Ethical considerations}

This work relies on publicly released datasets (\textsc{PsyDefConv} and 
\textsc{ESConv}) collected under informed consent and Institutional Review Board (IRB) oversight, with no 
new human data collection. Synthetic utterances simulating psychological distress 
are used exclusively for model training; clinical validity is not guaranteed, and 
expert review is required before any real-world deployment. The system is a 
research prototype and must not be used as a diagnostic tool.

\section*{Acknowledgments}
We would like to thank the organizers of the BioNLP 2026 PsyDefDetect shared task and acknowledge the \textsc{PsyDefConv} dataset as the foundation of this work.

\newpage
\bibliography{custom}

@INPROCEEDINGS{lin2018focallossdenseobject,
  author={Lin, Tsung-Yi and Goyal, Priya and Girshick, Ross and He, Kaiming and Dollár, Piotr},
  booktitle={2017 IEEE International Conference on Computer Vision (ICCV)}, 
  title={Focal Loss for Dense Object Detection}, 
  year={2017},
  volume={},
  number={},
  pages={2999-3007},
  doi={10.1109/ICCV.2017.324}}

@inproceedings{zhu2018texygenbenchmarkingplatformtext,
author = {Zhu, Yaoming and Lu, Sidi and Zheng, Lei and Guo, Jiaxian and Zhang, Weinan and Wang, Jun and Yu, Yong},
title = {Texygen: A Benchmarking Platform for Text Generation Models},
year = {2018},
isbn = {9781450356572},
publisher = {Association for Computing Machinery},
address = {New York, NY, USA},
url = {https://doi.org/10.1145/3209978.3210080},
doi = {10.1145/3209978.3210080},
booktitle = {The 41st International ACM SIGIR Conference on Research \& Development in Information Retrieval},
pages = {1097–1100},
numpages = {4},
location = {Ann Arbor, MI, USA},
series = {SIGIR '18}
}

@misc{kiela2020supervisedmultimodalbitransformersclassifying,
      title={Supervised Multimodal Bitransformers for Classifying Images and Text}, 
      author={Douwe Kiela and Suvrat Bhooshan and Hamed Firooz and Ethan Perez and Davide Testuggine},
      year={2020},
      eprint={1909.02950},
      archivePrefix={arXiv},
      primaryClass={cs.CL},
      url={https://arxiv.org/abs/1909.02950}, 
}

@article{Cohen1960ACO,
  title={A Coefficient of Agreement for Nominal Scales},
  author={Jacob Cohen},
  journal={Educational and Psychological Measurement},
  year={1960},
  volume={20},
  pages={37 - 46},
  url={https://api.semanticscholar.org/CorpusID:15926286}
}

@inproceedings{brown2020languagemodelsfewshotlearners,
author = {Brown, Tom B. and Mann, Benjamin and Ryder, Nick and Subbiah, Melanie and Kaplan, Jared and Dhariwal, Prafulla and Neelakantan, Arvind and Shyam, Pranav and Sastry, Girish and Askell, Amanda and Agarwal, Sandhini and Herbert-Voss, Ariel and Krueger, Gretchen and Henighan, Tom and Child, Rewon and Ramesh, Aditya and Ziegler, Daniel M. and Wu, Jeffrey and Winter, Clemens and Hesse, Christopher and Chen, Mark and Sigler, Eric and Litwin, Mateusz and Gray, Scott and Chess, Benjamin and Clark, Jack and Berner, Christopher and McCandlish, Sam and Radford, Alec and Sutskever, Ilya and Amodei, Dario},
title = {Language models are few-shot learners},
year = {2020},
isbn = {9781713829546},
publisher = {Curran Associates Inc.},
address = {Red Hook, NY, USA},
booktitle = {Proceedings of the 34th International Conference on Neural Information Processing Systems},
articleno = {159},
numpages = {25},
location = {Vancouver, BC, Canada},
series = {NIPS '20}
}

@inproceedings{
loshchilov2019decoupledweightdecayregularization,
title={Decoupled Weight Decay Regularization},
author={Ilya Loshchilov and Frank Hutter},
booktitle={International Conference on Learning Representations},
year={2019},
url={https://openreview.net/forum?id=Bkg6RiCqY7},
}

@inproceedings{kumar2021dataaugmentationusingpretrained,
    title = "Data Augmentation using Pre-trained Transformer Models",
    author = "Kumar, Varun  and
      Choudhary, Ashutosh  and
      Cho, Eunah",
    editor = "Campbell, William M.  and
      Waibel, Alex  and
      Hakkani-Tur, Dilek  and
      Hazen, Timothy J.  and
      Kilgour, Kevin  and
      Cho, Eunah  and
      Kumar, Varun  and
      Glaude, Hadrien",
    booktitle = "Proceedings of the 2nd Workshop on Life-long Learning for Spoken Language Systems",
    month = dec,
    year = "2020",
    address = "Suzhou, China",
    publisher = "Association for Computational Linguistics",
    url = "https://aclanthology.org/2020.lifelongnlp-1.3/",
    doi = "10.18653/v1/2020.lifelongnlp-1.3",
    pages = "18--26"
}

@inproceedings{wei-zou-2019-eda,
  title     = {{EDA}: Easy Data Augmentation Techniques for Boosting Performance on Text Classification Tasks},
  author    = {Wei, Jason and Zou, Kai},
  editor    = {Inui, Kentaro and Jiang, Jing and Ng, Vincent and Wan, Xiaojun},
  booktitle = {Proceedings of the 2019 Conference on Empirical Methods in Natural Language Processing and the 9th International Joint Conference on Natural Language Processing (EMNLP-IJCNLP)},
  month     = nov,
  year      = {2019},
  address   = {Hong Kong, China},
  publisher = {Association for Computational Linguistics},
  pages     = {6382--6388},
  doi       = {10.18653/v1/D19-1670},
  url       = {https://aclanthology.org/D19-1670/}
}

@article{anabytavor2019datadeeplearningrescue,
    title = "Do Not Have Enough Data? {Deep} Learning to the Rescue!",
    author = "Anaby-Tavor, Ateret and
              Carmeli, Boaz and
              Goldbraich, Esther and
              Kantor, Amir and
              Kour, George and
              Shlomov, Segev and
              Tepper, Naama and
              Zwerdling, Naama",
    journal = "Proceedings of the AAAI Conference on Artificial Intelligence",
    volume = "34",
    number = "05",
    pages = "7383--7390",
    year = "2020",
    month = apr,
    doi = "10.1609/aaai.v34i05.6233",
    url = "https://ojs.aaai.org/index.php/AAAI/article/view/6233",
}

@inproceedings{liu-etal-2021-towards,
  title     = {Towards Emotional Support Dialog Systems},
  author    = {Liu, Siyang and Zheng, Chujie and Demasi, Orianna and Sabour, Sahand and Li, Yu and Yu, Zhou and Jiang, Yong and Huang, Minlie},
  booktitle = {Proceedings of ACL-IJCNLP 2021},
  year      = {2021},
  publisher = {Association for Computational Linguistics},
  pages     = {3469--3483},
  doi       = {10.18653/v1/2021.acl-long.269},
  url       = {https://aclanthology.org/2021.acl-long.269/}
}

@article{bond83,
  title   = {An Empirical Study of Self-Rated Defense Style},
  author  = {Bond, Michael and Gardner, Susan and Christian, J. and Sigal, J.},
  journal = {Archives of General Psychiatry},
  year    = {1983},
  volume  = {40},
  pages   = {333--338}
}

@inproceedings{ji2022mentalbert,
    title = "{M}ental{BERT}: Publicly Available Pretrained Language Models for Mental Healthcare",
    author = "Ji, Shaoxiong  and
      Zhang, Tianlin  and
      Ansari, Luna  and
      Fu, Jie  and
      Tiwari, Prayag  and
      Cambria, Erik",
    editor = "Calzolari, Nicoletta  and
      B{\'e}chet, Fr{\'e}d{\'e}ric  and
      Blache, Philippe  and
      Choukri, Khalid  and
      Cieri, Christopher  and
      Declerck, Thierry  and
      Goggi, Sara  and
      Isahara, Hitoshi  and
      Maegaard, Bente  and
      Mariani, Joseph  and
      Mazo, H{\'e}l{\`e}ne  and
      Odijk, Jan  and
      Piperidis, Stelios",
    booktitle = "Proceedings of the Thirteenth Language Resources and Evaluation Conference",
    month = jun,
    year = "2022",
    address = "Marseille, France",
    publisher = "European Language Resources Association",
    url = "https://aclanthology.org/2022.lrec-1.778/",
    pages = "7184--7190"
}

@article{diGiuseppePerry2021,
  title   = {The Hierarchy of Defense Mechanisms: Assessing Defensive Functioning with the Defense Mechanisms Rating Scales {Q}-Sort},
  author  = {Di Giuseppe, Mariagrazia and Perry, J. Christopher},
  journal = {Frontiers in Psychology},
  year    = {2021},
  volume  = {12},
  pages   = {718440},
  doi     = {10.3389/fpsyg.2021.718440}
}

@article{Ji_2023,
  title   = {Survey of Hallucination in Natural Language Generation},
  author  = {Ji, Ziwei and Lee, Nayeon and Frieske, Rita and Yu, Tiezheng and Su, Dan and Xu, Yan and Ishii, Etsuko and Bang, Ye Jin and Madotto, Andrea and Fung, Pascale},
  journal = {ACM Computing Surveys},
  year    = {2023},
  volume  = {55},
  number  = {12},
  pages   = {1--38},
  doi     = {10.1145/3571730}
}

@article{cramer1987development,
  title   = {The Development of Defense Mechanisms},
  author  = {Cramer, Phebe},
  journal = {Journal of Personality},
  year    = {1987},
  volume  = {55},
  number  = {4},
  pages   = {597--614},
  doi     = {10.1111/j.1467-6494.1987.tb00454.x}
}

@article{vaillant1994ego,
  title   = {Ego Mechanisms of Defense and Personality Psychopathology},
  author  = {Vaillant, George E.},
  journal = {Journal of Abnormal Psychology},
  year    = {1994},
  volume  = {103},
  number  = {1},
  pages   = {44--50}
}

@inproceedings{na-etal-2026-psydefdetect, 
title = "Overview of the PsyDefDetect Shared Task at BioNLP 2026: Detecting Levels of Psychological Defense Mechanisms in Supportive Conversations", 
author = "Na, Hongbin and Wang, Zimu and Chen, Zhaoming and Hua, Yining and Gao, Rena and Yang, Kailai and Chen, Ling and Wang, Wei and Ji, Shaoxiong and Torous, John and Ananiadou, Sophia", 
booktitle = "Proceedings of the 25th Workshop on Biomedical Language Processing", 
month = jul, 
year = "2026", 
address = "San Diego, USA", 
publisher = "Association for Computational Linguistics", 
}

@inproceedings{na-etal-2025-survey, 
title = "A Survey of Large Language Models in Psychotherapy: Current Landscape and Future Directions", 
author = "Na, Hongbin and Hua, Yining and Wang, Zimu and Shen, Tao and Yu, Beibei and Wang, Lilin and Wang, Wei and Torous, John and Chen, Ling", 
booktitle = "Findings of the Association for Computational Linguistics: ACL 2025", 
month = jul, 
year = "2025", 
address = "Vienna, Austria", 
publisher = "Association for Computational Linguistics", 
url = "https://aclanthology.org/2025.findings-acl.385/",
doi = "10.18653/v1/2025.findings-acl.385", 
pages = "7362--7376", 
}

@inproceedings{na-etal-2026-psydefconv, 
title = "You Never Know a Person, You Only Know Their Defenses: Detecting Levels of Psychological Defense Mechanisms in Supportive Conversations", 
author = "Na, Hongbin and Wang, Zimu and Chen, Zhaoming and Zhou, Peilin and Hua, Yining and Zhou, Grace Ziqi and Zhang, Haiyang and Shen, Tao and Wang, Wei and Torous, John and Ji, Shaoxiong and Chen, Ling", 
booktitle = "Findings of the Association for Computational Linguistics: ACL 2026", 
month = jul, 
year = "2026", 
address = "San Diego, USA", 
publisher = "Association for Computational Linguistics", 
}

\newpage
\appendix
\section{Appendix}
\label{sec:appendix}

\subsection{Dataset Analysis}
\label{sec:appendix_dataset}

We conduct an exploratory analysis of \textsc{PsyDefConv} to characterize its
structural and temporal properties. Class imbalance is also reported
in \cite{na-etal-2026-psydefconv}.

\begin{figure}[H]
  \centering
  \includegraphics[width=\columnwidth]{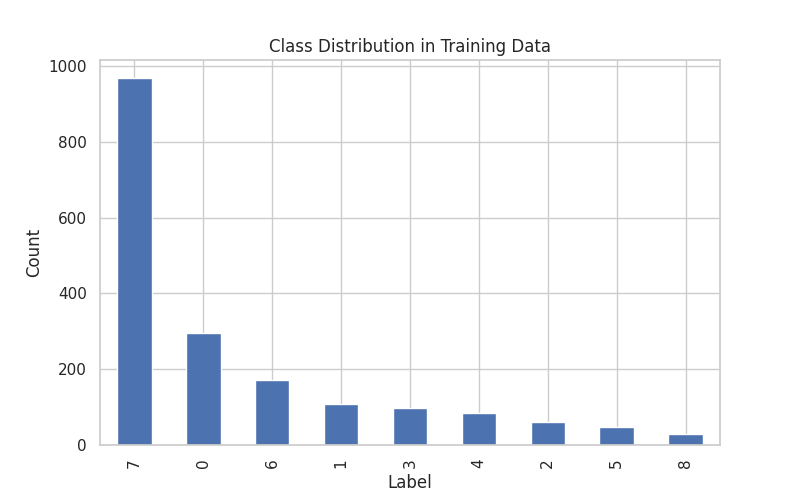}
  \caption{Class distribution across defense levels in the development set.
  Level 7 (High-Adaptive) dominates with 968 instances; Level 8 (Needs
  Info) contains only 28, motivating soft-balancing augmentation.}
  \label{fig:class_distribution}
\end{figure}

\paragraph{Temporal Volatility of Defense States.}

\begin{figure}[H]
  \includegraphics[width=\columnwidth]{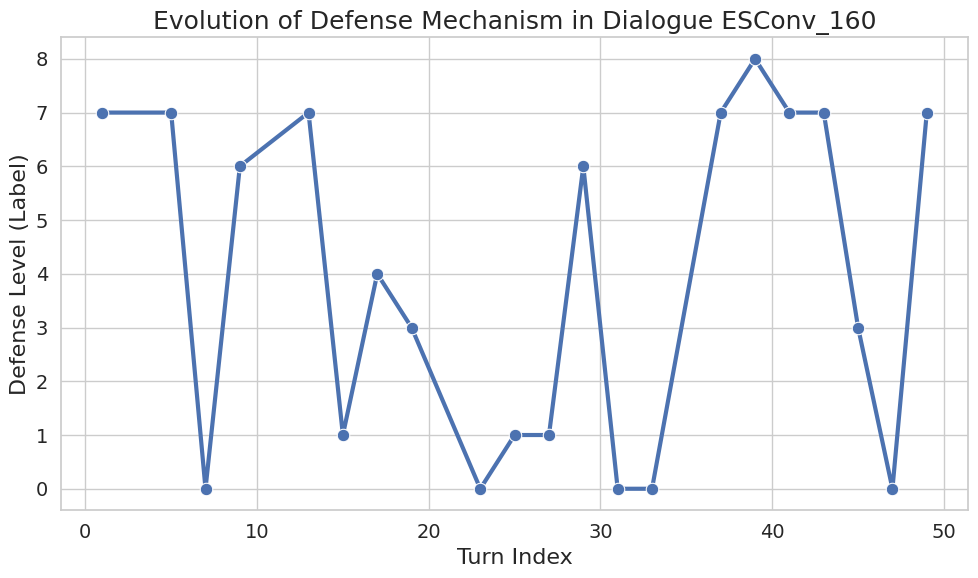}
  \caption{Defense level trajectory across turns in dialogue ESConv\_016.
  Frequent transitions, including abrupt shifts between Level 7 and
  Level 0, indicate that defense states are temporally unstable and
  cannot be modeled from isolated turns.}
  \label{fig:defense_evolution}
\end{figure}

Defense levels are not stable within a dialogue
(Fig.~\ref{fig:defense_evolution}). Frequent transitions across levels,
including abrupt shifts between adaptive and disavowal patterns within
a single session, indicate that classification cannot rely on static
turn-level features alone and must account for broader discourse context.

\paragraph{Defense Volatility: Magnitude and Speed of Change.}

\begin{figure}[t]
  \centering
  \includegraphics[width=\columnwidth]{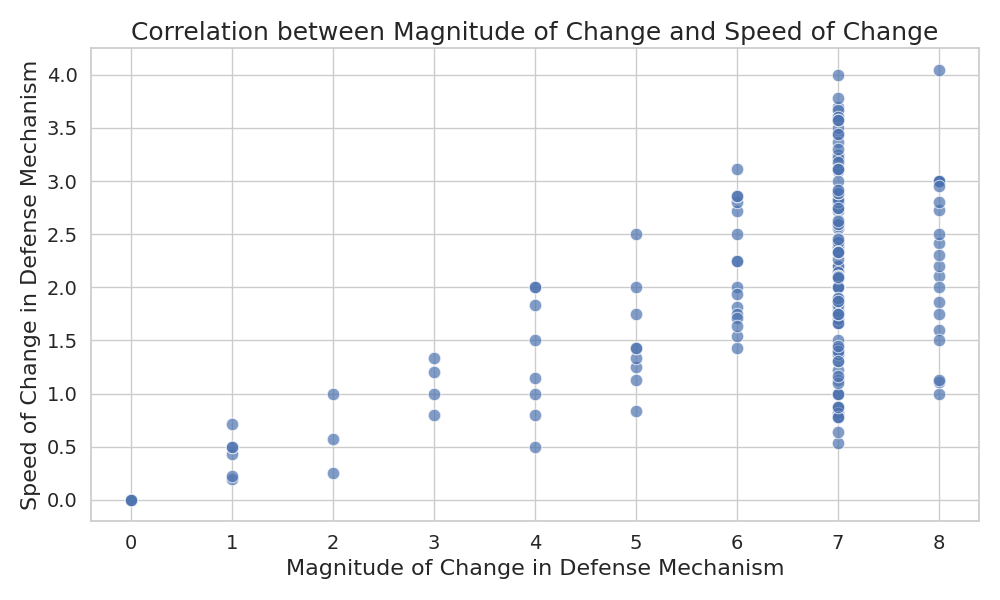}
  \caption{Correlation between magnitude and speed of defense level
  change across dialogue turns. Larger shifts in defense level tend to
  occur more rapidly, suggesting that defensive transitions are abrupt
  rather than gradual, a property that static turn-level classifiers
  are structurally unable to capture.}
  \label{fig:correlation_magnitude_speed}
\end{figure}

The scatter plot (Fig.~\ref{fig:correlation_magnitude_speed}) shows a
positive correlation between the magnitude and speed of defense-level
transitions: larger shifts in defensive functioning tend to occur over
fewer turns. This is a key empirical finding, it implies that when
a seeker's defense changes, it changes quickly and dramatically, rather
than gradually. This property motivates dialogue-level or
sequential modeling as a future direction, as turn-level classifiers
observe only the outcome of a transition, not its dynamics.

\paragraph{Disclosure Dynamics.}

\begin{figure}[t]
  \includegraphics[width=\columnwidth]{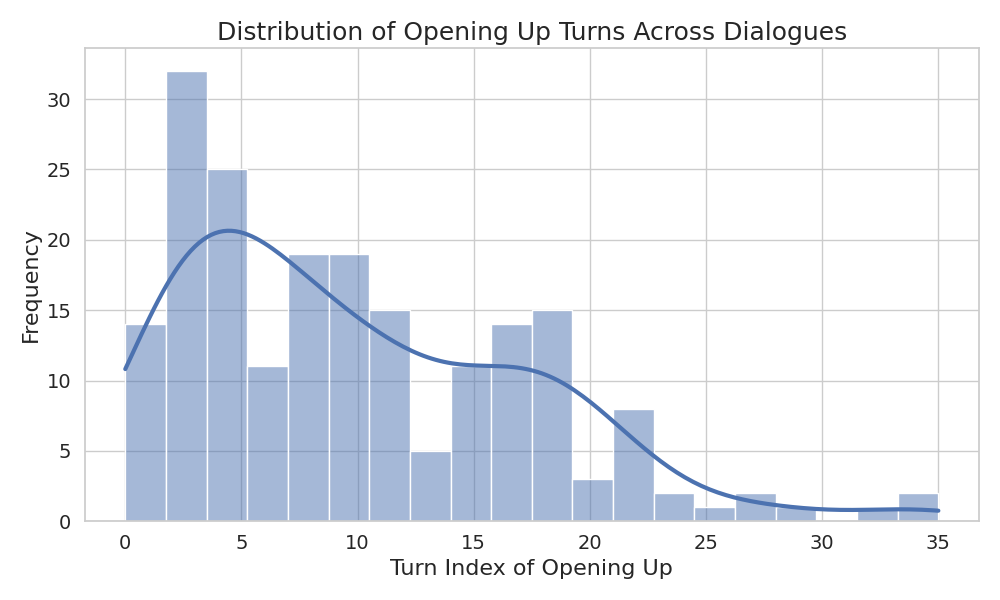}
  \caption{Distribution of turns at which seekers exhibit increased
  openness. The modal opening-up turn is 3--4, indicating early
  disclosure before defensive consolidation.}
  \label{fig:opening_up_turn_distribution}
\end{figure}

\begin{figure}[t]
  \includegraphics[width=\columnwidth]{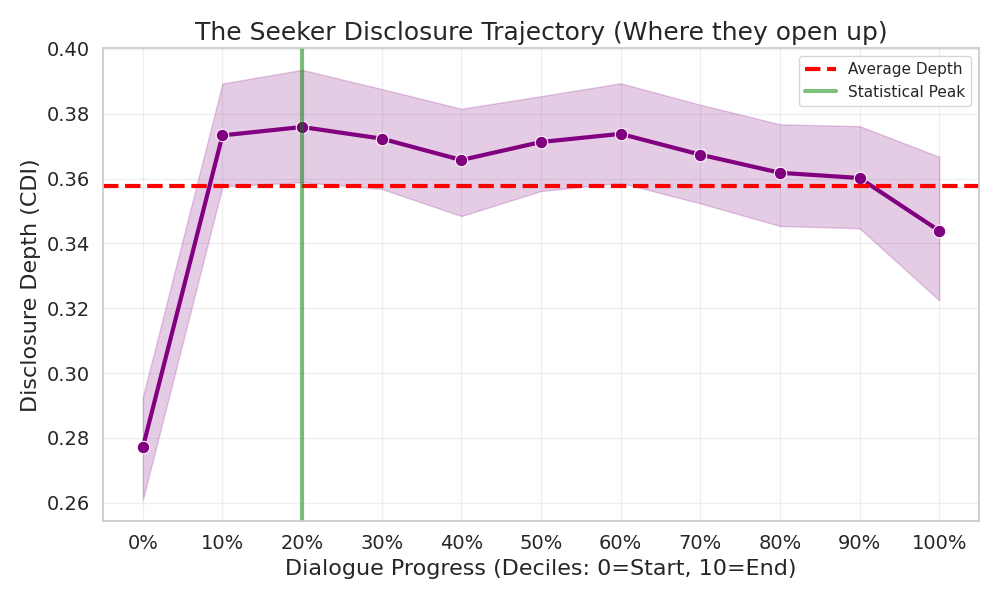}
  \caption{Composite Disclosure Index (CDI) across normalized dialogue
  progression. Disclosure peaks at the 10--20\% mark then stabilizes,
  suggesting defensive activation intensifies after initial vulnerability.}
  \label{fig:seeker_disclosure_trajectory}
\end{figure}

Analysis of the CDI (Fig.~\ref{fig:seeker_disclosure_trajectory}) reveals
a consistent disclosure peak around the 10-20\% mark, followed by
gradual stabilization. The opening-up distribution
(Fig.~\ref{fig:opening_up_turn_distribution}) confirms that seekers tend
to disclose early (modal turn $\approx$ 3-4), suggesting that defensive
activation intensifies \textit{after} initial vulnerability rather than
preceding it.

\paragraph{Response Latency as a Defensive Signal.}

\begin{figure}[t]
  \includegraphics[width=\columnwidth]{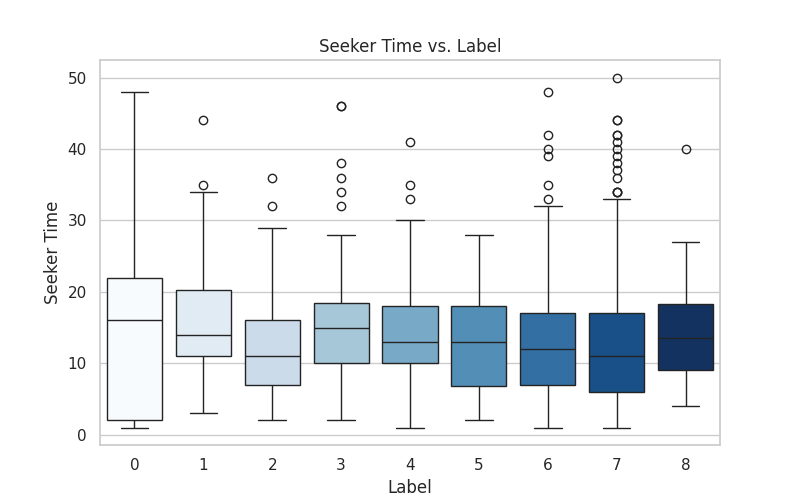}
  \caption{Seeker response time per defense label. Label 0 (No Defense)
  shows notably higher latency variance; defense-related labels cluster
  in tighter distributions, supporting temporal features as auxiliary
  classification signals.}
  \label{fig:seeker_time_vs_label}
\end{figure}

Seeker response time varies across defense levels
(Fig.~\ref{fig:seeker_time_vs_label}). Label 0 (No Defense) shows higher
latency variance, while defense-related responses cluster more tightly.
This supports the use of temporal features as auxiliary signals.

\paragraph{Corpus Structure.}

\begin{figure}[H]
  \includegraphics[width=\columnwidth]{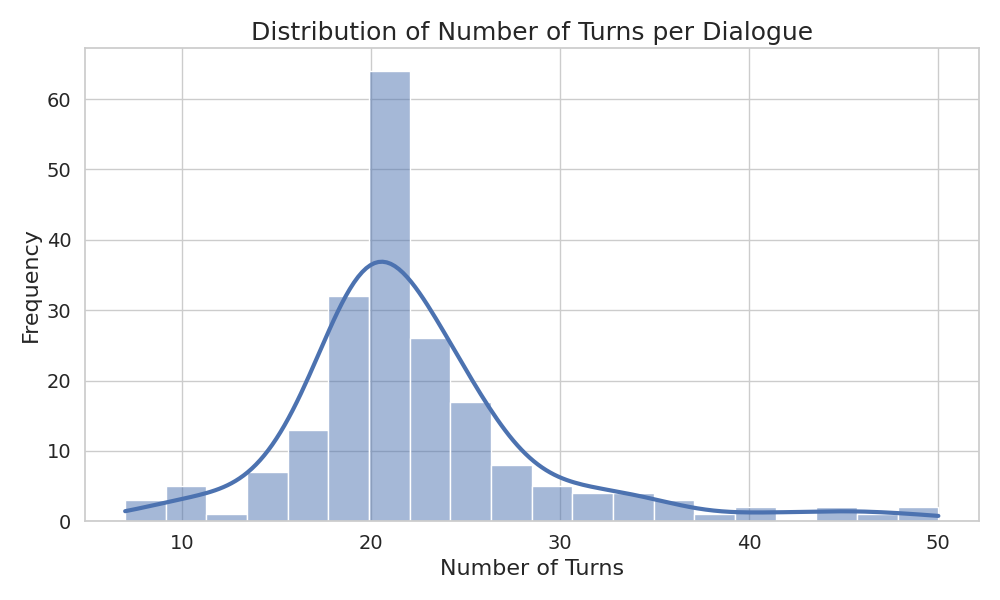}
  \caption{Distribution of number of turns per dialogue. Dialogues
  average approximately 20 turns, confirming the multi-turn nature
  of the classification task.}
  \label{fig:turns_per_dialogue}
\end{figure}

\begin{figure}[H]
  \includegraphics[width=\columnwidth]{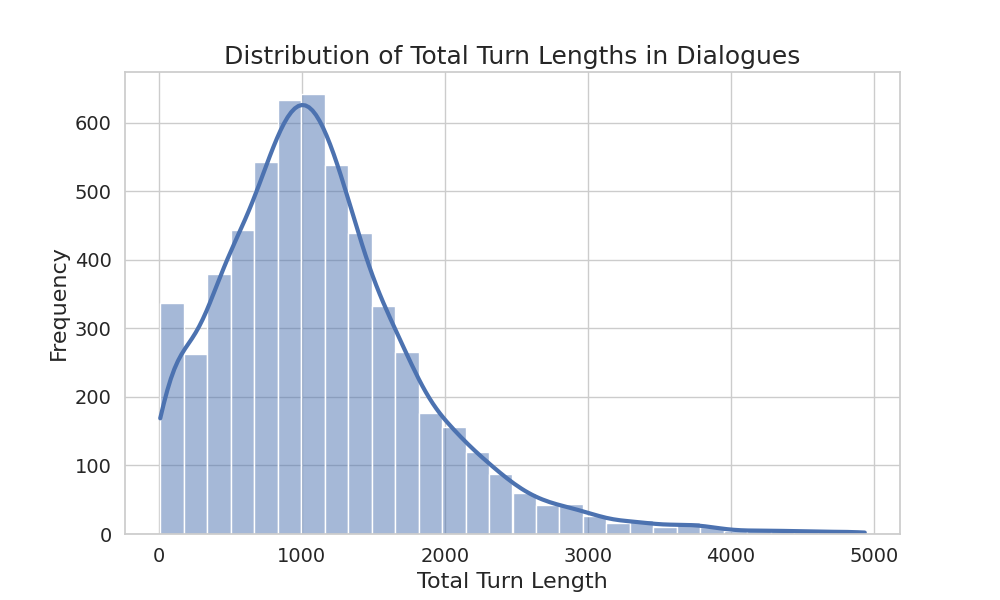}
  \caption{Distribution of total turn lengths per dialogue. The
  right-skewed distribution peaks around 1,000 tokens, with a long
  tail of extended sessions up to 5,000 tokens.}
  \label{fig:turn_length_distribution}
\end{figure}

Dialogues average 20 turns and 1,000 tokens in total turn length
(Figs.~\ref{fig:turns_per_dialogue} and \ref{fig:turn_length_distribution}),
confirming the multi-turn nature of the task and the need for
context-aware modeling beyond single utterances.

\subsection{Extended Result Analysis}
\label{sec:appendix_results}

\paragraph{Per-Class Metrics.}

\begin{figure}[!h]
  \includegraphics[width=\columnwidth]{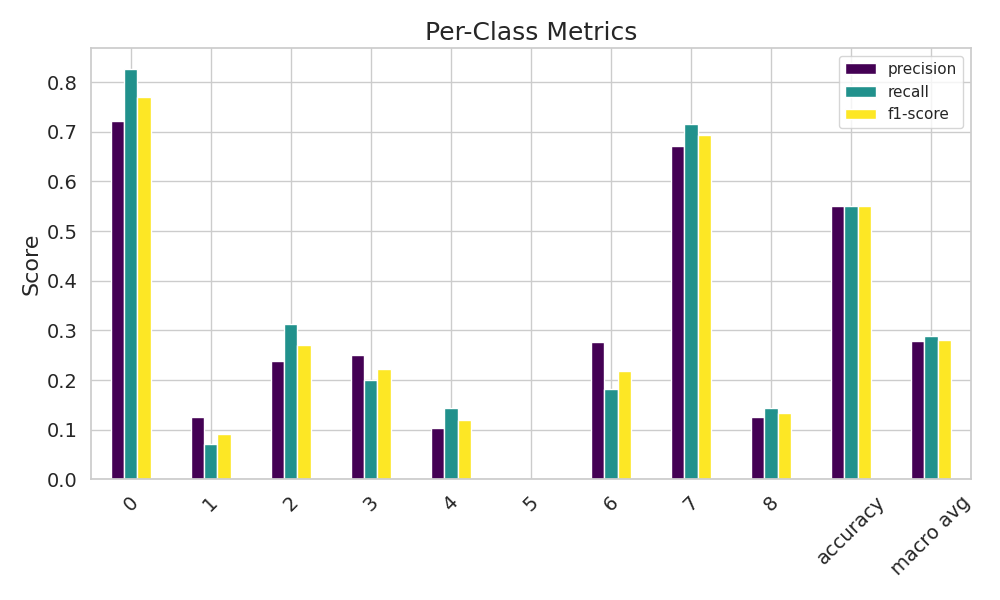}
  \caption{Per-class Precision, Recall, and F1 under the best setting
  ($\times$2). Labels 1, 4, 5, and 8 remain below F1 = 0.15; Label 5
  (Neurotic) achieves zero precision and recall, consistent with its
  severe underrepresentation (13 dev instances).}
  \label{fig:perclass_app}
\end{figure}

Labels 1, 4, 5, and 8 each yield F1 below 0.15. Label 5 (Neurotic)
is never predicted, consistent with only 13 dev instances and chronic
underrepresentation across all augmentation scales.

\paragraph{DMRS Mechanism Activation Patterns.}

\begin{figure}[!h]
  \includegraphics[width=\columnwidth]{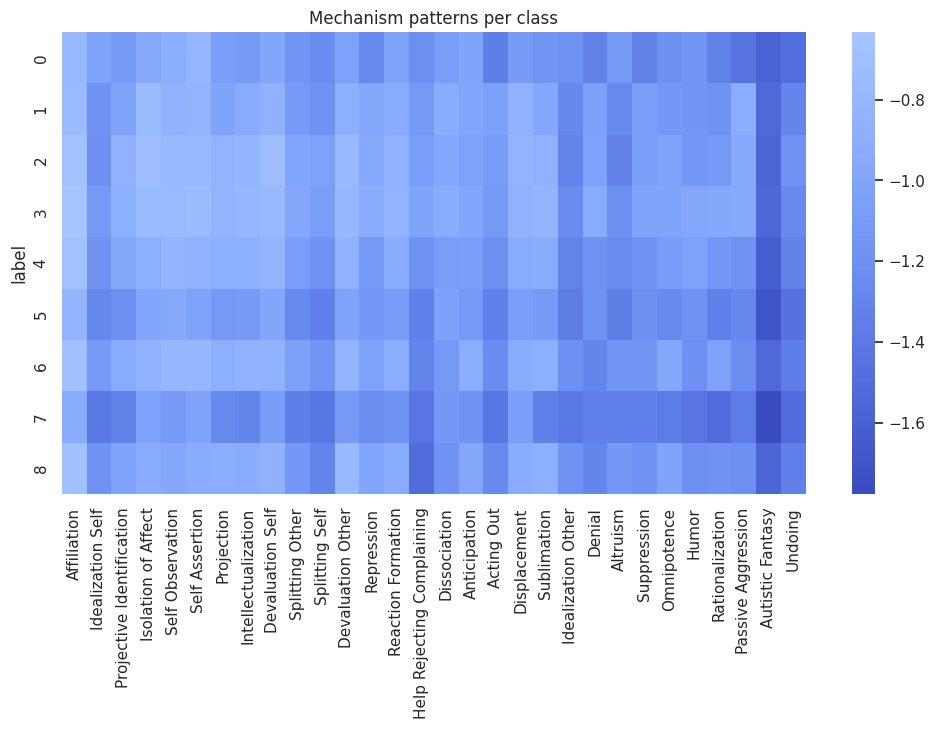}
  \caption{Mean NLI-inferred DMRS mechanism activation per defense class
  (log-entailment scores). All values are negative due to log-probability
  scaling. Differential gradients on \textit{Autistic Fantasy},
  \textit{Undoing}, and \textit{Affiliation} provide discriminative signal
  for the hybrid fusion model despite uniformly low absolute scores.}
  \label{fig:mechanism_app}
\end{figure}

Activation values are uniformly low (range: -0.8 to -1.7), reflecting
the implicit nature of defensive language. Despite this, differential
patterns across classes, particularly on \textit{Autistic Fantasy},
\textit{Undoing}, and \textit{Affiliation}, confirm that the Defense
Profile carries discriminative signal that complements the contextual
encoder in the fusion architecture.

\paragraph{Misclassification Patterns.}

\begin{figure}[H]
  \includegraphics[width=\columnwidth]{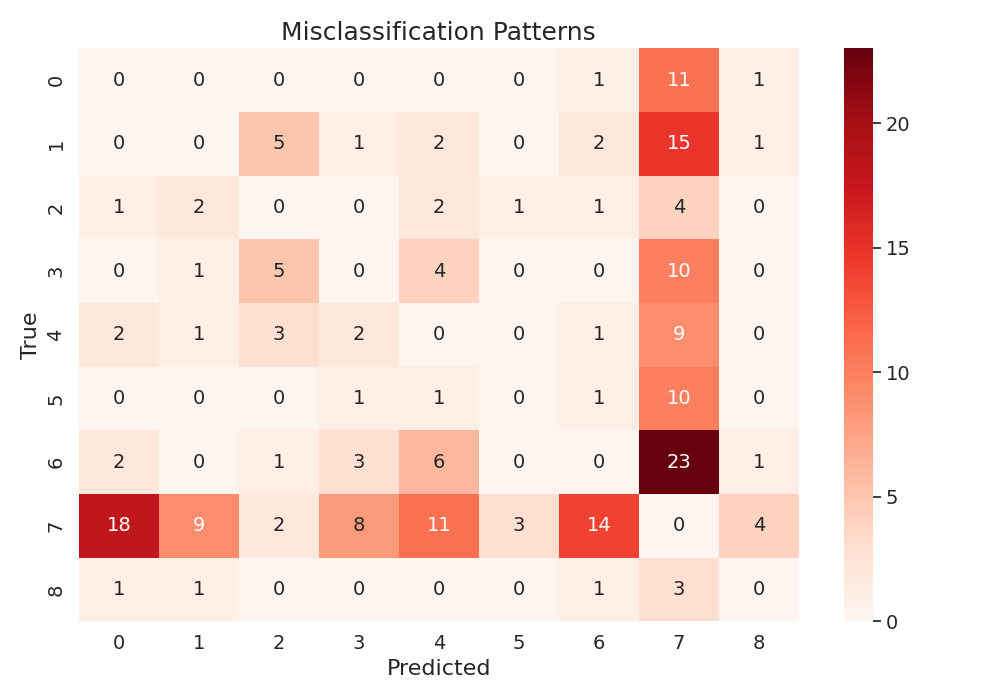}
  \caption{Off-diagonal misclassification counts (best setting, $\times$2).
  Label 7 acts as a prediction sink across all classes. The Label 6
  $\to$ 7 confusion (23 errors) is the largest single off-diagonal cell,
  reflecting clinical proximity between obsessional and high-adaptive defenses.}
  \label{fig:misclassification}
\end{figure}

The off-diagonal error analysis (Fig.~\ref{fig:misclassification}) reveals
that Label 7 is the dominant prediction sink: it absorbs the largest share
of errors from every other class, accounting for 11, 15, 10, 9, 10, 23,
and 3 misclassified samples from Labels 0-8 respectively. This is not
random confusion but a systematic bias toward the majority class.
Notably, Label 6 (Obsessional) is misclassified as Label 7 in 23 of 44
cases (52\%), suggesting high clinical proximity between obsessional and
high-adaptive functioning, a distinction that requires deeper contextual
modeling to resolve.

\newpage
\onecolumn
\subsection{Prompt Template for Synthetic Data Generation}
\label{appendix:prompt}

We employ a theory-driven prompting strategy to generate synthetic utterances conditioned on stressors, dialogue history, and clinical defense mechanisms. The template used for generation is shown below:








\paragraph{Llama3 Prompt}

\noindent
\begin{lstlisting}
prompt = f"""
    ### TASK: Generate Synthetic Psychological Defense Examples
    You are simulating a seeker in a mental health support chat.
    
    ### CONTEXTUAL GROUNDING:
    STRESSOR: {stressor}
    DIALOGUE HISTORY:
    {history}

    ### DEFENSE TO SIMULATE:
    Mechanism: {mechanism_name} (Level {level})
    Definition: {definition}
    Pattern: {pattern_description}

    ### REFERENCE STYLE (Few-Shot):
    1. "{example_1}"
    2. "{example_2}"
    3. "{example_3}"

    ### GOAL:
    Generate 5 NEW seeker utterances for the NEXT TURN using the {mechanism_name} defense.
    Ensure they follow the history and react to the stressor.

    ### OUTPUT FORMAT:
    1 string.
    No explanation, no markdown, no code fences.
    """
\end{lstlisting}

\subsection{Prompt Template for Stressor Identification}






\paragraph{Llama3 Prompt}

\noindent
\begin{lstlisting}
prompt = f"""
    ### TASK: Clinical Stressor Identification
    Identify the "Salient Stressor" causing psychological conflict in the Target Utterance.

    ### DIALOGUE CONTEXT:
    {history}

    ### TARGET UTTERANCE:
    "{target_turn}"

    ### OUTPUT FORMAT:
    1. Stressor Category: (e.g., Interpersonal Conflict, Self-Esteem Threat, External Crisis)
    2. Description: (One sentence explaining the threat)
    """
\end{lstlisting}

\end{document}